\documentclass[twoside,11pt]{article}

\usepackage{jmlr2e}
\usepackage[utf8]{inputenc} 
\usepackage[T1]{fontenc}    
\usepackage{hyperref}       
\usepackage{url}            
\usepackage{booktabs}       
\usepackage{amsfonts}       
\usepackage{nicefrac}       
\usepackage{microtype}      
\usepackage{graphicx}
\usepackage{graphics}
\usepackage{xcolor}
\usepackage[frozencache=true,cachedir=minted-cache]{minted}
\usepackage{subfigure}
\usepackage{lscape}
\usepackage{multirow}
\usepackage{longtable}
\usepackage{here}
\usepackage{siunitx}[retain-zero-uncertainty=true]

\definecolor{LightGray}{gray}{0.9}

\usepackage{lastpage}
\jmlrheading{23}{2022}{1-\pageref{LastPage}}{1/22; Revised
9/22}{10/22}{22-0017}{Takuma Seno and Michita Imai}
\ShortHeadings{title}{Seno and Imai}

\ShortHeadings{d3rlpy: An Offline Deep Reinforcement Learning Library}{Seno and Imai}
\firstpageno{1}

\begin{document}

\title{d3rlpy: An Offline Deep Reinforcement Learning Library}

\author{\name Takuma Seno \email seno@ailab.ics.keio.ac.jp \\
       \addr Keio University \\
       Kanagawa, Japan \\
       Sony AI\\
       Tokyo, Japan
       \AND
       \name Michita Imai \email michita@ailab.ics.keio.ac.jp \\
       \addr Keio University\\
       Kanagawa, Japan}

\editor{Alexandre Gramfort}

\maketitle

\begin{abstract}
    In this paper, we introduce \texttt{d3rlpy}, an open-sourced offline deep reinforcement learning (RL) library for Python.
    \texttt{d3rlpy} supports a set of offline deep RL algorithms as well as off-policy online algorithms via a fully documented plug-and-play API.
    To address a reproducibility issue, we conduct a large-scale benchmark with D4RL and Atari 2600 dataset to ensure implementation quality and provide experimental scripts and full tables of results.
    The \texttt{d3rlpy} source code can be found on GitHub: \url{https://github.com/takuseno/d3rlpy}.
\end{abstract}

\begin{keywords}
  offline reinforcement learning, deep reinforcement learning, reproducibility, open source software, pytorch
\end{keywords}

\section{Introduction}
Deep reinforcement learning (RL) has been led to significant advancements in numerous domains such as gaming~\citep{sophy} and robotics~\citep{quadrupedal}.
While RL algorithms have a potential to solve complex tasks, active data collection is a major challenge especially for environments where interaction is expensive.
Offline RL~\citep{offline}, where algorithms find a good policy within a previously collected static dataset, has been considered as a solution to this problem.

Although recent offline deep RL papers are published with author-provided implementations, they are scattered across different repositories and do not provide standardized interfaces, which makes it difficult for researchers to incorporate the algorithms into their projects.
It is also crucial for researchers to have an access to faithfully benchmarked implementations to deal with a reproducibility problem~\citep{rl_matters}.

In this paper, we introduce \texttt{Data-Driven Deep Reinforcement Learning library for Python} (\texttt{d3rlpy}), an offline deep RL library for Python.
\texttt{d3rlpy} provides a set of off-policy offline and online RL algorithms built with \texttt{PyTorch}~\citep{pytorch}.
API of all implemented algorithms is fully documented, plug-and-play and standardized so that users can easily start experiments with \texttt{d3rlpy}.
To solve the reproducibility issue in offline RL, a large-scale faithful benchmark is conducted with \texttt{d3rlpy}.

\section{Related work}
The choice of API design determines user experience, which has to balance the tradeoff between ease of use and flexibility.
\texttt{KerasRL}~\citep{kerasrl} and \texttt{Stable-Baselines3}~\citep{sb3} provide deep RL algorithms with plug-and-play API and extensive documentations.
\texttt{Tensorforce}~\citep{tensorforce}, \texttt{MushroomRL}~\citep{mushroomrl} and \texttt{Tianshou}~\citep{tianshou} provide moduralized deep RL components that allow users to conduct custom experiments.
\texttt{SaLinA}~\citep{salina} provides a general framework for decision-making agents where users can implement scalable algorithms on top of it.
To encourage a broader RL community to start offline RL research, \texttt{d3rlpy} is designed to be the first library that privdes fully documented plug-and-play API for offline RL experiments.

From the reproducibility perspective, \texttt{ChainerRL}~\citep{chainerrl} and \texttt{Tonic}~\citep{tonic} provide many deep RL algorithms with faithful reproduction results.
\texttt{Dopamine}~\citep{dopamine} is also the widely used implementation in the community, which focuses on making DQN-variants~\citep{rainbow} available for researchers.
\texttt{d3rlpy} is also the first library accompanied by a number of offline RL algorithms and the extensive benchmark results in this research field.

Integrating the fully documented plug-and-play API and a number of faithfully benchmarked offline RL algorithms altogether, it is difficult for the existing libraries to instantly achieve the same offline RL research experience as \texttt{d3rlpy}.

\section{Design of d3rlpy}

In this section, the library design of \texttt{d3rlpy} is described.

\subsection{Library interface}
\begin{minted}
[
frame=lines,
bgcolor=LightGray,
breaklines,
fontsize=\scriptsize
]{python}
import d3rlpy

# prepare MDPDataset object
dataset, env = d3rlpy.datasets.get_dataset("hopper-medium-v0")
# prpare Algorithm object
sac = d3rlpy.algos.SAC(actor_learning_rate=3e-4, use_gpu=0)
# start offline training
sac.fit(
    dataset,
    n_steps=500000,
    eval_episodes=dataset.episodes,
    scorers={"environment": d3rlpy.metrics.evaluate_on_environment(env)},
)
# seamlessly start online training
sac.fit_online(env, n_steps=500000)
\end{minted}
\texttt{d3rlpy} provides scikit-learn-styled API~\citep{scikit_learn} to make the use of this library as easy as possible.
In terms of the library design, there are two main differences from the existing libraries.
First, \texttt{d3rlpy} has an interface for offline RL training, which takes a dedicated RL dataset component, \texttt{MDPDataset} described in Section 3.2.
Second, all methods for training such as \texttt{fit} and \texttt{fit\_online} are implemented in \texttt{Algorithm} components to make \texttt{d3rlpy} as plug-and-play as possible.
For the plug-and-play user experience, neural network architectures are automatically selected from MLP and the convolutional model~\citep{dqn} depending on observation, which allows users to start training without composing neural network models unless using customized architectures described in Section 3.2.
These design choices are expected to lower the bar to start using this library.

Since \texttt{d3rlpy} supports both offline and online training, the seamless transition from offline training to online fine-tuning is realized.
Fine-tuning policies trained offline is demanded, but is still a challenging problem~\citep{awac,iql}.
This seamless transition supports the further research by allowing RL researchers to easily conduct fine-tuning experiments.

\subsection{Components}
\begin{figure}[t]
\begin{center}
    \includegraphics[width=0.8\columnwidth]{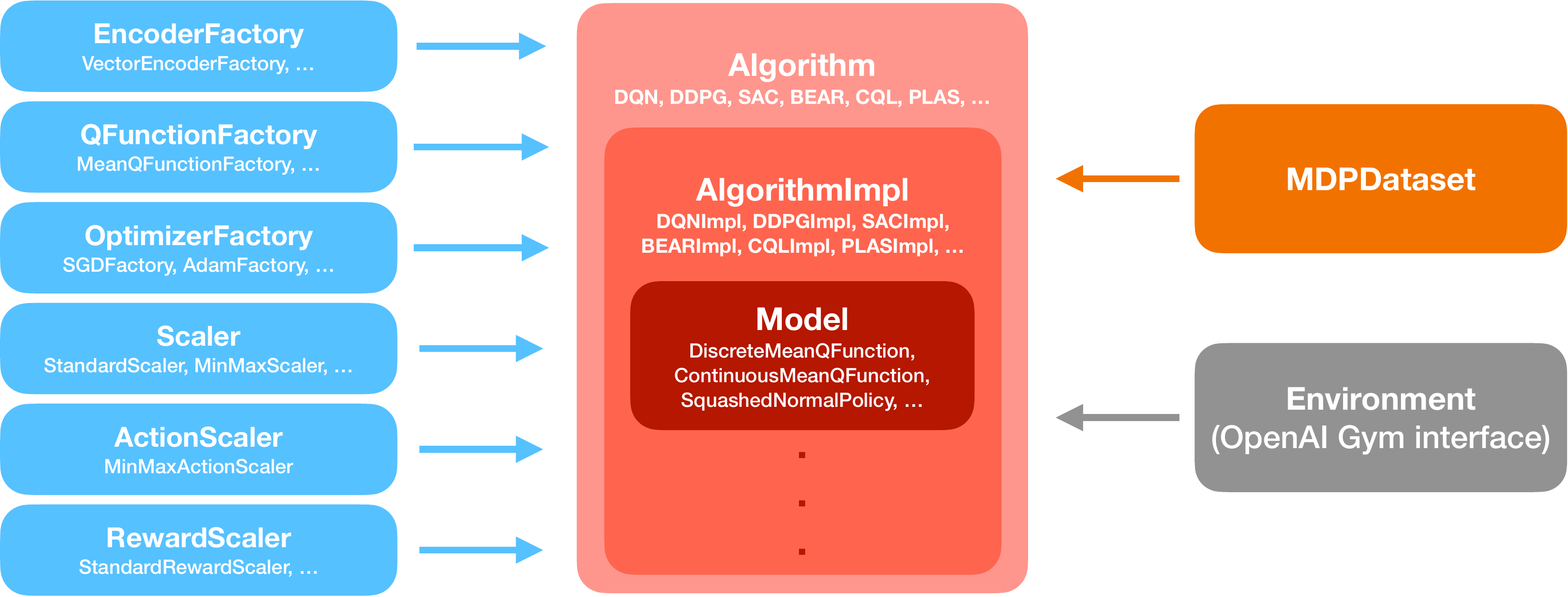}
\end{center}
    \caption{The illustration of module components in \texttt{d3rlpy}. \texttt{MDPDataset} and OpenAI Gym-styled environment can be used to train policies.}
  \label{fig:design}
\end{figure}

We highlight main components provided by \texttt{d3rlpy}.
Figure~\ref{fig:design} depicts module components in \texttt{d3rlpy}.
All of these components provide the standardized API.
The full documentation including extensive tutorials of the library is available at \url{https://d3rlpy.readthedocs.io}.

\textbf{Algorithm.}
\texttt{Algorithm} components provide the offline and online training methods described in Section 3.1. \texttt{Algorithm} is implemented in a hierarchical design that internally instantiates \texttt{AlgorithmImpl}.
This hierarchy is to provide high-level user-friendly API such as \texttt{fit} method for \texttt{Algorithm} and low-level API such as \texttt{update\_actor} and \texttt{update\_critic} for \texttt{AlgorithmImpl}.
The main motivation of this hierarchical API system is to increase module resusability of the algoirthms when the new algorithm only requires high-level changes.
For example, \textit{delayed policy update} of TD3, which updates policy parameters every two gradient steps, can be implemented by adjusting frequency of \texttt{update\_actor} method calls in the high-level module without changing the low-level logics.

\textbf{MDPDataset.}
\texttt{MDPDataset} provides the standardized offline RL dataset interface.
Users can build their own dataset by using logged data consisting with numpy arrays~\citep{numpy} of \texttt{observations}, \texttt{actions}, \texttt{rewards}, \texttt{terminals} and optionally \texttt{timeouts}~\citep{timeout}.
The popular benchmark datasets such as D4RL and Atari 2600 datasets are also provided by \texttt{d3rlpy.datasets} package that converts them into \texttt{MDPDataset} object.
In addition, \texttt{d3rlpy} supports automatic data collection by giving OpenAI Gym~\citep{gym} style environment, which exports collected data as \texttt{MDPDataset} object.
For diverse sets of dataeset creation, the data collection can be performed with and without parameter updates.

\textbf{EncoderFactory.}
User-defined custom neural network models are supported via \texttt{EncoderFactory} components.
Users can define all function approximators to train in \texttt{d3rlpy} by building their own \texttt{EncoderFactory} components.
This flexibility allows the use of the complex architectures~\citep{resnet} and experiments with partially pretrained models~\citep{imagenet_rl}.

\textbf{QFunctionFactory.}
\texttt{d3rlpy} provides \texttt{QFunctionFactory} components that allow users to use distributional Q-functions: Quantile Regression~\citep{qr_dqn} and Implicit Quantile Network~\citep{iqn}.
The distributional Q-functions dramatically improve performance by capturing variance of returns.
Unlike conventional RL libraries that implement distributional Q-functions as DQN-variants, \texttt{d3rlpy} enables users to use them with all implemented algorithms, which reduces complexity to support algorithmic-variants such as QR-DQN~\citep{qr_dqn} and a discrete version of CQL~\citep{cql}.

\textbf{Scaler, ActionScaler and RewardScaler.}
By exploiting static dataset in offline RL training, \texttt{d3rlpy} provides various preprocessing and postprocessing methods through \texttt{Scaler}, \texttt{ActionScaler} and \texttt{RewardScaler} components.
For observation preprocessing, \textit{normalization}, \textit{standardization} and \textit{pixel} are available.
The observation \texttt{standardization} has been shown to improve policy performance in offline RL setting~\citep{td3_plus_bc}.
Regarding action preprocessing, \textit{normalization} is available and action output from a trained policy is denormalized to original scale as postprocessing.
Lastly, reward preprocessing supports \textit{normalization}, \textit{standardization}, \textit{clip} and \textit{constant multiplication}.

\section{Large-scale benchmark}
To address the reproducibility problem~\citep{rl_matters}, the implemented algorithms\footnote[1]{\texttt{d3rlpy} implements NFQ~\citep{nfq}, DQN~\citep{dqn}, Double DQN~\citep{double_dqn}, DDPG~\citep{ddpg}, TD3~\citep{td3}, SAC~\citep{sac}, BCQ~\citep{bcq}, BEAR~\citep{bear}, CQL~\citep{cql}, AWAC~\citep{awac}, CRR~\citep{crr}, PLAS~\citep{plas}, PLAS+P~\citep{plas}, TD3+BC~\citep{td3_plus_bc} and IQL~\citep{iql}.} are faithfully benchmarked with D4RL~\citep{d4rl} and Atari 2600 datasets~\citep{optimistic}.
The full Python scripts used in this benchmark are also included in our source code~\footnote[2]{\url{https://github.com/takuseno/d3rlpy/tree/master/reproductions}}, which allows users to conduct additional benchmark experiments.
Full tables of benchmark results are reported in Appendix A, Appendix B and Appendix C.
The all logged metrics are released in our GitHub repository~\footnote[3]{\url{https://github.com/takuseno/d3rlpy-benchmarks}}.

\section{Conclusion}
In this paper, we introduced an offline deep reinforcement learning library, \texttt{d3rlpy}.
\texttt{d3rlpy} provides a set of offline and online RL algorithms through the standardized plug-and-play API.
The large-scale faithful benchmark was conducted to address the reproducibility issue.


\acks{This work is supported by Information-technology Promotion Agency, Japan (IPA), Exploratory IT Human Resources Project (MITOU Program) in the fiscal year 2020.
We would like to express our genuine gratitude for the contributions made by the voluntary contributors.
We would also like to thank our users who have provided constructive feedback and insights.}

\vskip 0.2in


\appendix

\section*{Appendix A. Benchmark Results: D4RL}
\label{sec:appendix_d4rl}
We evaluated the implemented algorithms on the open-sourced D4RL benchmark of OpenAI Gym MuJoCo tasks~\citep{gym,d4rl}.
We followed the experimental procedure described in \cite{d4rl}.
We trained each algorithm for 500K gradient steps and evaluated every 5000 steps to collect evaluation performance in environments for 10 episodes.
Table~\ref{tab:params} shows hyperparameters used in benchmarking.
We used the same hyperparameters as the ones previously reported in previous papers or recommended in author-provided repositories.
We used discount factor of $0.99$, target update rate of 5e-3 and an Adam optimizer~\citep{adam} across all algorithms.
The default architecture was MLP with hidden layers of [256, 256] unless we explicitly address it.
We repeated all experiments with 10 random seeds.

Table~\ref{tab:benchmark_final} shows results of the benchmark in normalized scale~\citep{d4rl}.
Table~\ref{tab:sac_ref}, \ref{tab:awac_ref}, \ref{tab:bcq_ref}, \ref{tab:bear_ref}, \ref{tab:cql_ref}, \ref{tab:iql_ref}, \ref{tab:plas_ref}, \ref{tab:plas_p_ref} and \ref{tab:td3_plus_bc_ref} show side-by-side comparisons with reference scores.
CRR is not included in this side-by-side comparison because CRR has not been benchmarked with D4RL and an author-provided implementation is not publically available.
Considering the fact that standard deviations for some algorithms were not reported by the authors and many algorithms in offline RL research were originally benchmarked with only 3 or 4 random seeds, we believe that performance discrepancy is trivial.

\begin{scriptsize}
\begin{longtable}[c]{c|l|l}
    \hline
    Algorithm                                    & Hyperparameter            & Value \\
    \hline
    \multirow{3}{*}{SAC~\citep{sac}}             & Critic learning rate      & 3e-4 \\
                                                 & Actor learning rate       & 3e-4 \\
                                                 & Mini-batch size           & 256 \\
    \hline
    \multirow{7}{*}{AWAC~\citep{awac}}           & Critic learning rate      & 3e-4 \\
                                                 & Actor learning rate       & 3e-4 \\
                                                 & Mini-batch size           & 1024 \\
                                                 & $\lambda$~\citep{awac}    & 1 \\
                                                 & Actor hidden units        & [256, 256, 256, 256] \\
                                                 & Actor weight decay        & 1e-4 \\
                                                 & Critic hidden units       & [256, 256, 256, 256] \\
    \hline
    \multirow{12}{*}{BCQ~\citep{bcq}}            & Critic learning rate      & 1e-3 \\
                                                 & Actor learning rate       & 1e-3 \\
                                                 & VAE learning rate         & 1e-3 \\
                                                 & Mini-batch size           & 100 \\
                                                 & $\lambda$~\citep{bcq}     & 0.75 \\
                                                 & Critic hidden units       & [400, 300] \\
                                                 & Actor hidden units        & [400, 300] \\
                                                 & VAE encoder hidden units  & [750, 750] \\
                                                 & VAE decoder hidden units  & [750, 750] \\
                                                 & VAE latent size           & $2 \times |A|$ \\
                                                 & Perturbation range        & 0.05 \\
                                                 & Action samples            & 100 \\
    \hline
    \multirow{15}{*}{BEAR~\citep{bear}\footnotemark[4]}& Critic learning rate      & 3e-4 \\
                                                 & Actor learning rate       & 1e-4 \\
                                                 & VAE learning rate         & 3e-4 \\
                                                 & $\alpha$ learning rate    & 1e-3 \\
                                                 & Mini-batch size           & 256 \\
                                                 & VAE encoder hidden units  & [750, 750] \\
                                                 & VAE decoder hidden units  & [750, 750] \\
                                                 & VAE latent size           & $2 \times |A|$ \\
                                                 & MMD $\sigma$              & 20 \\
                                                 & MMD kernel                & laplacian (gaussian for HalfCheetah) \\
                                                 & MMD action samples        & 4 \\
                                                 & $\alpha$ threshold        & 0.05 \\
                                                 & Target action samples     & 10 \\
                                                 & Evaluation action samples & 100 \\
                                                 & Pretraining steps         & 40000 \\
    \hline
    \multirow{7}{*}{CQL~\citep{cql}\footnotemark[5]}& Critic learning rate      & 3e-4 \\
                                                 & Actor learning rate       & 1e-4 \\
                                                 & Fixed $\alpha$            & 5 (10 for medium datasets) \\
                                                 & Mini-batch size           & 256 \\
                                                 & Critic hidden units       & [256, 256, 256] \\
                                                 & Actor hidden units        & [256, 256, 256] \\
                                                 & Action samples            & 10 \\
    \hline
    \multirow{6}{*}{CRR~\citep{crr}}             & Critic learning rate      & 3e-4 \\
                                                 & Actor learning rate       & 3e-4 \\
                                                 & Mini-batch size           & 256 \\
                                                 & Action samples            & 4 \\
                                                 & Advantage type            & mean \\
                                                 & Weight type               & binary \\
    \hline
    \multirow{7}{*}{IQL~\citep{iql}}             & Critic learning rate      & 3e-4 \\
                                                 & Actor learning rate       & 3e-4 \\
                                                 & V-function learning rate  & 3e-4 \\
                                                 & Mini-batch size           & 256 \\
                                                 & Expectile                 & 0.7 \\
                                                 & Inverse temperature       & 3.0 \\
                                                 & Actor learning rate scheduler & Cosine \\
    \hline
    \multirow{11}{*}{PLAS~\citep{plas}}          & Critic learning rate      & 1e-3 \\
                                                 & Actor learning rate       & 1e-4 \\
                                                 & VAE learning rate         & 1e-4 \\
                                                 & Mini-batch size           & 100 \\
                                                 & Critic hidden units       & [400, 300] \\
                                                 & Actor hidden units        & [400, 300] \\
                                                 & VAE encoder hidden units  & [750, 750] ([128, 128] for medium-replay) \\
                                                 & VAE decoder hidden units  & [750, 750] ([128, 128] for medium-replay) \\
                                                 & $\lambda$~\citep{bcq}     & 1.0 \\
                                                 & VAE latent size           & $2 \times |A|$ \\
                                                 & VAE pretraining steps     & 500000 \\
    \hline
    \multirow{12}{*}{PLAS+P~\citep{plas}}        & Critic learning rate      & 1e-3 \\
                                                 & Actor learning rate       & 1e-4 \\
                                                 & VAE learning rate         & 1e-4 \\
                                                 & Mini-batch size           & 100 \\
                                                 & Critic hidden units       & [400, 300] \\
                                                 & Actor hidden units        & [400, 300] \\
                                                 & VAE encoder hidden units  & [750, 750] ([128, 128] for medium-replay) \\
                                                 & VAE decoder hidden units  & [750, 750] ([128, 128] for medium-replay) \\
                                                 & $\lambda$~\citep{bcq}     & 1.0 \\
                                                 & VAE latent size           & $2 \times |A|$ \\
                                                 & VAE pretraining steps     & 500000 \\
                                                 & Perturbation range        & Appendix D in \cite{plas} \\
    \hline
    \multirow{8}{*}{TD3+BC~\citep{td3_plus_bc}}  & Critic learning rate      & 3e-4 \\
                                                 & Actor learning rate       & 3e-4 \\
                                                 & Mini-batch size           & 256 \\
                                                 & Policy noise              & 0.2 \\
                                                 & Policy noise clipping     & (-0.5, 0.5) \\
                                                 & Policy update frequency   & 2 \\
                                                 & $\alpha$~\citep{td3_plus_bc} & 2.5 \\
                                                 & Observation preprocess    & standardization \\
    \hline
     \caption{Hyperparameters for D4RL.}
     \label{tab:params}
\end{longtable}
\end{scriptsize}
\footnotetext[4]{\url{https://github.com/Farama-Foundation/d4rl_evaluations}}
\footnotetext[5]{\url{https://github.com/aviralkumar2907/CQL}}

\begin{landscape}
\begin{table*}[p]
  \tiny
  \centering
    \begin{tabular}{|l||S[table-format=2.1(2.1)]||S[table-format=2.1(2.1)]|S[table-format=3.1(2.1)]|S[table-format=2.1(2.1)]|S[table-format=3.1(2.1)]|S[table-format=2.1(2.1)]|S[table-format=3.1(2.1)]|S[table-format=3.1(2.1)]|S[table-format=3.1(2.1)]|S[table-format=3.1(2.1)]|}
     \hline
      \textbf{Dataset}               & \textbf{SAC} & \textbf{AWAC}  & \textbf{BCQ}  & \textbf{BEAR} & \textbf{CQL}  & \textbf{CRR}  & \textbf{IQL}  & \textbf{PLAS}  & \textbf{PLAS+P} & \textbf{TD3+BC} \\
	 \hline
	 \hline
      halfcheetah-random-v0         &  30.2 \pm 1.9 & 15.2 \pm 1.3   & 2.3 \pm 0.0   & 2.3 \pm 0.0   & 29.7 \pm 1.5  & 18.9 \pm 7.0  & 14.4 \pm 2.3  & 26.9 \pm 1.4   & 27.3 \pm 2.3    & 11.4 \pm 1.5  \\
	 \hline
     walker2d-random-v0             & 2.5 \pm 1.7   & 4.3 \pm 1.9    & 4.2 \pm 1.5   & 4.6 \pm 1.4   & 2.0 \pm 2.8   & 2.3 \pm 1.9   & 5.8 \pm 0.3   & 6.5 \pm 7.2    & 4.5 \pm 5.0     & 2.1 \pm 2.4 \\
	 \hline
     hopper-random-v0               & 1.1 \pm 0.6   & 11.1 \pm 0.1   & 10.5 \pm 0.2  & 10.1 \pm 0.2  & 10.8 \pm 0.1  & 10.9 \pm 1.3  & 11.1 \pm 0.1  & 10.4 \pm 0.3   & 12.5 \pm 1.5    & 10.9 \pm 0.1 \\
	 \hline
	 \hline
     halfcheetah-medium-v0          & 30.1 \pm 10.6 & 41.9 \pm 0.4   & 40.1 \pm 0.5  & 36.6 \pm 0.9  & 41.7 \pm 0.2  & 41.9 \pm 0.4  & 41.1 \pm 0.2  & 39.8 \pm 0.4   & 42.0 \pm 0.6    & 42.4 \pm 1.5 \\
	 \hline
     walker2d-medium-v0             & 2.1 \pm 5.3   & 65.6 \pm 11.3  & 47.7 \pm 7.8  & 57.4 \pm 10.5 & 77.8 \pm 5.1  & 43.3 \pm 17.5 & 59.7 \pm 8.9  & 33.0 \pm 9.7   & 66.1 \pm 6.8    & 76.7 \pm 3.2 \\
	 \hline
     hopper-medium-v0               & 1.2 \pm 0.8   & 40.7 \pm 20.3  & 52.5 \pm 22.8 & 34.8 \pm 8.0  & 50.1 \pm 19.9 & 26.1 \pm 32.3 & 31.1 \pm 0.3  & 58.1 \pm 23.7  & 53.6 \pm 24.0   & 95.9 \pm 12.1 \\
	 \hline
	 \hline
     halfcheetah-medium-replay-v0   & 38.1 \pm 7.3  & 42.4 \pm 1.4   & 39.0 \pm 1.9  & 37.1 \pm 2.0  & 43.6 \pm 3.0  & 42.0 \pm 0.4  & 40.9 \pm 0.9  & 43.9 \pm 0.4   & 44.8 \pm 1.0    & 42.8 \pm 2.1 \\
	 \hline
     walker2d-medium-replay-v0      & 4.5 \pm 3.8   & 22.5 \pm 7.0   & 15.2 \pm 4.6  & 13.6 \pm 3.1  & 20.5 \pm 4.2  & 26.7 \pm 4.6  & 14.3 \pm 4.4  & 20.9 \pm 13.3  & 9.2 \pm 10.4    & 26.4 \pm 5.7 \\
	 \hline
     hopper-medium-replay-v0        & 8.5 \pm 11.8  & 34.0 \pm 4.1   & 16.1 \pm 7.7  & 27.8 \pm 6.3  & 31.1 \pm 3.4  & 13.3 \pm 14.0 & 38.1 \pm 5.3  & 17.5 \pm 13.3  & 20.0 \pm 26.2   & 31.4 \pm 8.0 \\
	 \hline
	 \hline
     halfcheetah-medium-expert-v0   & 0.0 \pm 2.3   & 16.1 \pm 4.2   & 60.1 \pm 11.1 & 45.4 \pm 7.4  & 9.0 \pm 2.8   & 8.4 \pm 1.7   & 55.2 \pm 7.9  & 81.2 \pm 9.6   & 81.6 \pm 8.1    & 89.2 \pm 6.8 \\
	 \hline
     walker2d-medium-expert-v0      & 1.7 \pm 2.9   & 7.9 \pm 22.0   & 43.6 \pm 14.0 & 59.1 \pm 9.6  & 54.7 \pm 37.3 & 41.1 \pm 12.3 & 92.3 \pm 22.3 & 93.5 \pm 9.1   & 86.3 \pm 10.4   & 91.0 \pm 14.4 \\
	 \hline
     hopper-medium-expert-v0        & 7.8 \pm 9.1   & 42.7 \pm 46.5  & 111.5 \pm 2.8 & 77.8 \pm 19.9 & 105.1 \pm 7.8 & 0.8 \pm 0.1   & 112.3 \pm 0.2 & 110.8 \pm 31.8 & 77.5 \pm 50.4   & 112.3 \pm 0.3 \\
	 \hline
  \end{tabular}
    \caption{Normalized scores and standard deviation collected with the final trained policy in D4RL. The scores and are averaged over 10 random seeds.}
  \label{tab:benchmark_final}
\end{table*}
\end{landscape}

\begin{table}[h]
  \centering
    \begin{tabular}{|l||S[table-format=2.1(2.1)]|r|}
     \hline
      \textbf{Dataset}               & {d3rlpy}       & reference \\
	 \hline
	 \hline
      halfcheetah-random-v0          & 30.2 \pm 1.9   & 30.5      \\
	 \hline
     walker2d-random-v0              & 2.5 \pm 1.7    & 4.1        \\
	 \hline
     hopper-random-v0                & 1.1 \pm 0.6    & 11.3        \\
	 \hline
	 \hline
     halfcheetah-medium-v0           & 30.1 \pm 10.6  & -4.3       \\
	 \hline
     walker2d-medium-v0              & 2.1 \pm 5.3    & 0.9        \\
	 \hline
     hopper-medium-v0                & 1.2 \pm 0.8    & 0.8        \\
	 \hline
	 \hline
     halfcheetah-medium-replay-v0    & 38.1 \pm 7.3   & -2.4       \\
	 \hline
     walker2d-medium-replay-v0       & 4.5 \pm 3.8    & 1.9        \\
	 \hline
     hopper-medium-replay-v0         & 8.5 \pm 11.8   & 3.5       \\
	 \hline
	 \hline
     halfcheetah-medium-expert-v0    & 0.0 \pm 2.3    & 1.8       \\
	 \hline
     walker2d-medium-expert-v0       & 1.7 \pm 2.9    & -0.1        \\
	 \hline
     hopper-medium-expert-v0         & 7.8 \pm 9.1    & 1.6        \\
	 \hline
  \end{tabular}
    \caption{Side-by-side comparison with reference normalized scores of SAC reported in \cite{d4rl}, which only provides mean scores.}
  \label{tab:sac_ref}
\end{table}

\begin{table}[h]
  \centering
    \begin{tabular}{|l||S[table-format=2.1(2.1)]|r|}
     \hline
      \textbf{Dataset}               & {d3rlpy}       & reference \\
	 \hline
	 \hline
      halfcheetah-random-v0          & 15.2 \pm 1.3   & 2.2      \\
	 \hline
     walker2d-random-v0              & 4.3 \pm 1.9    & 5.1        \\
	 \hline
     hopper-random-v0                & 11.1 \pm 0.1   & 9.6        \\
	 \hline
	 \hline
     halfcheetah-medium-v0           & 41.9 \pm 0.4   & 37.4       \\
	 \hline
     walker2d-medium-v0              & 65.6 \pm 11.3  & 30.1        \\
	 \hline
     hopper-medium-v0                & 40.7 \pm 20.3  & 72.0        \\
	 \hline
	 \hline
     halfcheetah-medium-replay-v0    & 42.4 \pm 1.4   & -       \\
	 \hline
     walker2d-medium-replay-v0       & 22.5 \pm 7.0   & -        \\
	 \hline
     hopper-medium-replay-v0         & 34.0 \pm 4.1   & -       \\
	 \hline
	 \hline
     halfcheetah-medium-expert-v0    & 16.1 \pm 4.2  & 36.8       \\
	 \hline
     walker2d-medium-expert-v0       & 7.9 \pm 22.0  & 42.7        \\
	 \hline
     hopper-medium-expert-v0         & 42.7 \pm 46.5 & 80.9        \\
	 \hline
  \end{tabular}
    \caption{Side-by-side comparison with reference normalized scores of AWAC reported in \cite{awac}, which only provides mean scores.}
  \label{tab:awac_ref}
\end{table}

\begin{table}[h]
  \centering
    \begin{tabular}{|l||S[table-format=3.1(2.1)]|r|}
     \hline
      \textbf{Dataset}              & {d3rlpy}      & reference \\
	 \hline
	 \hline
      halfcheetah-random-v0         & 2.3 \pm 0.0   & 2.2      \\
	 \hline
     walker2d-random-v0             & 4.2 \pm 1.5   & 4.9        \\
	 \hline
     hopper-random-v0               & 10.5 \pm 0.2  & 10.6        \\
	 \hline
	 \hline
     halfcheetah-medium-v0          & 40.1 \pm 0.5  & 40.7       \\
	 \hline
     walker2d-medium-v0             & 47.7 \pm 7.8  & 53.1        \\
	 \hline
     hopper-medium-v0               & 52.5 \pm 22.8 & 54.5        \\
	 \hline
	 \hline
     halfcheetah-medium-replay-v0   & 16.1 \pm 7.7  & 38.2       \\
	 \hline
     walker2d-medium-replay-v0      & 15.2 \pm 4.6  & 15.0        \\
	 \hline
     hopper-medium-replay-v0        & 16.1 \pm 7.7  & 33.1       \\
	 \hline
	 \hline
     halfcheetah-medium-expert-v0   & 60.1 \pm 11.1 & 64.7       \\
	 \hline
     walker2d-medium-expert-v0      & 43.6 \pm 14.0 & 57.5        \\
	 \hline
     hopper-medium-expert-v0        & 111.5 \pm 2.8 & 110.9        \\
	 \hline
  \end{tabular}
    \caption{Side-by-side comparison with reference normalized scores of BCQ reported in \cite{d4rl}, which only provides mean scores.}
  \label{tab:bcq_ref}
\end{table}

\begin{table}[h]
  \centering
    \begin{tabular}{|l||S[table-format=2.1(2.1)]|S[table-format=2.1(2.1)]|}
     \hline
      \textbf{Dataset}               & {d3rlpy}      & {reference} \\
	 \hline
	 \hline
      halfcheetah-random-v0          & 2.3 \pm 0.0   & 2.3 \pm 2.3      \\
	 \hline
     walker2d-random-v0              & 4.6 \pm 1.4   & 9.0 \pm 6.2        \\
	 \hline
     hopper-random-v0                & 10.1 \pm 0.2  & 10.0 \pm 0.7        \\
	 \hline
	 \hline
     halfcheetah-medium-v0           & 36.6 \pm 0.9  & 37.1 \pm 2.3       \\
	 \hline
     walker2d-medium-v0              & 57.4 \pm 10.5 & 56.1 \pm 8.5        \\
	 \hline
     hopper-medium-v0                & 34.8 \pm 8.0  & 30.8 \pm 0.9        \\
	 \hline
	 \hline
     halfcheetah-medium-replay-v0    & 37.1 \pm 2.0  & 36.2 \pm 5.6       \\
	 \hline
     walker2d-medium-replay-v0       & 13.6 \pm 3.1  & 13.7 \pm 2.1        \\
	 \hline
     hopper-medium-replay-v0         & 27.8 \pm 6.3  & 31.1 \pm 0.9       \\
	 \hline
	 \hline
     halfcheetah-medium-expert-v0    & 45.4 \pm 7.4  & 44.2 \pm 13.8       \\
	 \hline
     walker2d-medium-expert-v0       & 59.1 \pm 9.6  & 43.8 \pm 6.0        \\
	 \hline
     hopper-medium-expert-v0         & 77.8 \pm 19.9 & 67.3 \pm 32.5        \\
	 \hline
  \end{tabular}
    \caption{Side-by-side comparison with reference normalized scores of BEAR collected by executing the author-provided implementation to match the hyperparameters suggested in their GitHub page.}
  \label{tab:bear_ref}
\end{table}

\begin{table}[h]
  \centering
    \begin{tabular}{|l||S[table-format=3.1(2.1)]|S[table-format=3.1(2.1)]|}
     \hline
      \textbf{Dataset}               & {d3rlpy}       & {reference} \\
	 \hline
	 \hline
      halfcheetah-random-v0          & 29.7 \pm 1.5   & 28.5 \pm 2.4      \\
	 \hline
     walker2d-random-v0              & 2.0 \pm 2.8    & 1.2 \pm 2.6       \\
	 \hline
     hopper-random-v0                & 10.8 \pm 0.1   & 10.6 \pm 0.8        \\
	 \hline
	 \hline
     halfcheetah-medium-v0           & 41.7 \pm 0.2   & 38.8 \pm 2.5       \\
	 \hline
     walker2d-medium-v0              & 77.8 \pm 5.1   & 48.7 \pm 22.2        \\
	 \hline
     hopper-medium-v0                & 50.1 \pm 19.9  & 31.2 \pm 1.0        \\
	 \hline
	 \hline
     halfcheetah-medium-replay-v0    & 43.6 \pm 3.0   & 44.9 \pm 2.8       \\
	 \hline
     walker2d-medium-replay-v0       & 20.5 \pm 4.2   & 25.5 \pm 13.0        \\
	 \hline
     hopper-medium-replay-v0         & 31.1 \pm 3.4   & 30.1 \pm 2.2       \\
	 \hline
	 \hline
     halfcheetah-medium-expert-v0    & 9.0 \pm 2.8    & 11.3 \pm 4.9       \\
	 \hline
     walker2d-medium-expert-v0       & 54.7 \pm 37.3  & 75.4 \pm 52.8        \\
	 \hline
     hopper-medium-expert-v0         & 105.1 \pm 7.8  & 100.0 \pm 18.6        \\
	 \hline
  \end{tabular}
    \caption{Side-by-side comparison with reference normalized scores of CQL collected by executing the author-provided implementation to match the hyperparameters suggested in their GitHub page.}
  \label{tab:cql_ref}
\end{table}

\begin{table}[h]
  \centering
    \begin{tabular}{|l||S[table-format=3.1(2.1)]|S[table-format=3.1(2.1)]|}
     \hline
      \textbf{Dataset}               & {d3rlpy}       & {reference} \\
	 \hline
	 \hline
      halfcheetah-random-v0          & 14.4 \pm 2.3   & 12.2 \pm 3.4      \\
	 \hline
     walker2d-random-v0              & 5.8 \pm 0.3    & 5.7 \pm 0.1        \\
	 \hline
     hopper-random-v0                & 11.1 \pm 0.1   & 11.3 \pm 0.1        \\
	 \hline
	 \hline
     halfcheetah-medium-v0           & 41.1 \pm 0.2   & 41.0 \pm 0.4       \\
	 \hline
     walker2d-medium-v0              & 59.7 \pm 8.9   & 62.8 \pm 5.5        \\
	 \hline
     hopper-medium-v0                & 31.1 \pm 0.3   & 31.6 \pm 0.3        \\
	 \hline
	 \hline
     halfcheetah-medium-replay-v0    & 40.9 \pm 0.9   & 39.2 \pm 1.6       \\
	 \hline
     walker2d-medium-replay-v0       & 14.3 \pm 4.4   & 15.3 \pm 2.4        \\
	 \hline
     hopper-medium-replay-v0         & 38.1 \pm 5.3   & 39.1 \pm 2.7       \\
	 \hline
	 \hline
     halfcheetah-medium-expert-v0    & 55.2 \pm 7.9   & 54.3 \pm 2.2       \\
	 \hline
     walker2d-medium-expert-v0       & 92.3 \pm 22.3  & 101.1 \pm 7.4        \\
	 \hline
     hopper-medium-expert-v0         & 112.3 \pm 0.2  & 100.5 \pm 16.8        \\
	 \hline
  \end{tabular}
    \caption[Caption for IQL]{Side-by-side comparison with reference normalized scores of IQL collected by executing the author-provided\footnotemark[6] implementation because the scores for -v0 datasets were not reported in their original paper~\citep{iql}.}
  \label{tab:iql_ref}
\end{table}
\footnotetext[6]{\url{https://github.com/ikostrikov/implicit_q_learning}}

\begin{table}[h]
  \centering
    \begin{tabular}{|l||S[table-format=3.1(2.1)]|r|}
     \hline
      \textbf{Dataset}               & {d3rlpy}      & reference \\
	 \hline
	 \hline
      halfcheetah-random-v0          & 26.9 \pm 1.4   & 25.8      \\
	 \hline
     walker2d-random-v0              & 6.5 \pm 7.2    & 3.1        \\
	 \hline
     hopper-random-v0                & 10.4 \pm 0.3   & 10.5        \\
	 \hline
	 \hline
     halfcheetah-medium-v0           & 39.8 \pm 0.4   & 39.3       \\
	 \hline
     walker2d-medium-v0              & 33.0 \pm 9.7   & 44.6        \\
	 \hline
     hopper-medium-v0                & 58.1 \pm 23.7  & 32.9        \\
	 \hline
	 \hline
     halfcheetah-medium-replay-v0    & 38.1 \pm 7.3   & 43.9       \\
	 \hline
     walker2d-medium-replay-v0       & 20.9 \pm 13.3  & 30.2        \\
	 \hline
     hopper-medium-replay-v0         & 17.5 \pm 13.3  & 27.9       \\
	 \hline
	 \hline
     halfcheetah-medium-expert-v0    & 81.2 \pm 9.6   & 96.6       \\
	 \hline
     walker2d-medium-expert-v0       & 93.5 \pm 9.1   & 89.6        \\
	 \hline
     hopper-medium-expert-v0         & 100.8 \pm 31.8 & 110.0        \\
	 \hline
  \end{tabular}
    \caption{Side-by-side comparison with reference normalized scores of PLAS reported in \cite{plas}, which only provides mean scores.}
  \label{tab:plas_ref}
\end{table}

\begin{table}[h]
  \centering
    \begin{tabular}{|l||S[table-format=2.1(2.1)]|r|}
     \hline
      \textbf{Dataset}               & {d3rlpy}       & reference \\
	 \hline
	 \hline
      halfcheetah-random-v0          & 27.3 \pm 2.3   & 28.3      \\
	 \hline
     walker2d-random-v0              & 4.5 \pm 5.0    & 6.8        \\
	 \hline
     hopper-random-v0                & 12.5 \pm 1.5   & 13.3        \\
	 \hline
	 \hline
     halfcheetah-medium-v0           & 42.0 \pm 0.6   & 42.2       \\
	 \hline
     walker2d-medium-v0              & 66.1 \pm 6.8   & 66.9        \\
	 \hline
     hopper-medium-v0                & 53.6 \pm 24.0  & 36.9        \\
	 \hline
	 \hline
     halfcheetah-medium-replay-v0    & 44.8 \pm 1.0   & 45.7       \\
	 \hline
     walker2d-medium-replay-v0       & 9.2 \pm 10.4   & 14.3        \\
	 \hline
     hopper-medium-replay-v0         & 20.0 \pm 26.2  & 51.9       \\
	 \hline
	 \hline
     halfcheetah-medium-expert-v0    & 81.6 \pm 8.1   & 99.3       \\
	 \hline
     walker2d-medium-expert-v0       & 86.3 \pm 10.4  & 96.2        \\
	 \hline
     hopper-medium-expert-v0         & 77.5 \pm 50.4  & 94.7        \\
	 \hline
  \end{tabular}
    \caption{Side-by-side comparison with reference normalized scores of PLAS+P reported in \cite{plas}, which only provides mean scores.}
  \label{tab:plas_p_ref}
\end{table}

\begin{table}[h]
  \centering
    \begin{tabular}{|l||S[table-format=3.1(2.1)]|S[table-format=3.1(1.1)]|}
     \hline
      \textbf{Dataset}               & {d3rlpy}       & {reference} \\
	 \hline
	 \hline
      halfcheetah-random-v0          & 11.4 \pm 1.5   & 10.2 \pm 1.3      \\
	 \hline
     walker2d-random-v0              & 2.1 \pm 2.5    & 1.4 \pm 1.6        \\
	 \hline
     hopper-random-v0                & 10.9 \pm 0.1   & 11.0 \pm 0.1        \\
	 \hline
	 \hline
     halfcheetah-medium-v0           & 42.4 \pm 0.5   & 42.8 \pm 0.3       \\
	 \hline
     walker2d-medium-v0              & 76.7 \pm 3.2   & 79.7 \pm 1.8        \\
	 \hline
     hopper-medium-v0                & 95.9 \pm 12.1  & 99.5 \pm 1.0        \\
	 \hline
	 \hline
     halfcheetah-medium-replay-v0    & 42.8 \pm 2.1   & 43.3 \pm 0.5       \\
	 \hline
     walker2d-medium-replay-v0       & 26.4 \pm 5.7   & 25.2 \pm 5.1        \\
	 \hline
     hopper-medium-replay-v0         & 31.4 \pm 8.0   & 31.4 \pm 3.0       \\
	 \hline
	 \hline
     halfcheetah-medium-expert-v0    & 89.2 \pm 6.8   & 97.9 \pm 4.4       \\
	 \hline
     walker2d-medium-expert-v0       & 91.0 \pm 14.4  & 101.1 \pm 9.3        \\
	 \hline
     hopper-medium-expert-v0         & 112.3 \pm 0.3  & 112.2 \pm 0.2        \\
	 \hline
  \end{tabular}
    \caption{Side-by-side comparison with reference normalized scores of TD3+BC reported in \cite{td3_plus_bc}, which provides standard deviations as well as mean scores.}
  \label{tab:td3_plus_bc_ref}
\end{table}

\clearpage
\section*{Appendix B. Benchmark Results: Atari 2600}
\label{sec:appendix_atari}
We evaluated the implemented algorithms with open-sourced Atari 2600 datasets~\citep{optimistic}.
We followed the experimental procedure described in \cite{optimistic}.
We used 1\% portion of transitions (500K datapoints) and train each algorithm for 12.5M gradient steps and evaluate every 125K steps to collect evaluation performance in environments for 10 episodes.
Table~\ref{tab:params_atari} shows hyperparameters used in benchmarking.
We used the same hyperparameters for QR-DQN and CQL as the ones reported in \cite{cql}.
For NFQ and BCQ, the hyperparameters were chosen based on the QR-DQN setup for fair comparison because there are no benchmark results for them with publically available datasets.
We used discount factor of $0.99$, Adam optimizer~\citep{adam} and the convolutional neural network~\citep{dqn} across all algorithms.
Note that we configured BCQ with the Quantile Regression Q-function introduced in Section 3.2 to match the CQL setup.
In evaluation, we used $\epsilon$-greedy of $\epsilon=0.001$ and 25\% probability of sticky action~\citep{optimistic}.
We repeated all experiments with 10 random seeds.

Table~\ref{tab:benchmark_atari} shows the benchmark results, and Table~\ref{tab:qr_dqn_ref} and \ref{tab:discrete_cql_ref} show side-by-side comparisons with reference scores.
NFQ and BCQ are not included in the side-by-side comparisons because those are not evaluated with publically available datasets, an author-provided implementation for NFQ is not publically available, and the author-provided implementation\footnote[7]{\url{https://github.com/sfujim/BCQ}} for BCQ is not directly applicable for this evaluation.
Considering that the authors did not report standard deviations, we believe that performance discrepancy is trivial.

\begin{longtable}[c]{c|l|l}
    \hline
    Algorithm                  & Hyperparameter               & Value \\
    \hline
    \multirow{2}{*}{NFQ}       & Learning rate                & 5e-5 \\
                               & Mini-batch size              & 32 \\
    \hline
    \multirow{5}{*}{QR-DQN}    & Learning rate                & 5e-5 \\
                               & Mini-batch size              & 32 \\
                               & $\epsilon$~\citep{adam}      & 3.125e-4 \\
                               & Number of quantiles          & 200 \\
                               & Target update frequency      & 2000 \\
    \hline
    \multirow{7}{*}{BCQ}       & Learning rate                & 5e-5 \\
                               & Mini-batch size              & 32 \\
                               & $\epsilon$~\citep{adam}      & 3.125e-4 \\
                               & Number of quantiles          & 200 \\
                               & Target update frequency      & 2000 \\
                               & $\tau$~\citep{discrete_bcq}  & 0.3 \\
                               & Pre-activation regularization~\citep{discrete_bcq} & 1e-2 \\
    \hline
    \multirow{6}{*}{CQL}       & Learning rate                & 5e-5 \\
                               & Mini-batch size              & 32 \\
                               & $\epsilon$~\citep{adam}      & 3.125e-4 \\
                               & Number of quantiles          & 200 \\
                               & Target update frequency      & 2000 \\
                               & $\alpha$~\citep{cql}         & 4.0 \\
    \hline
     \caption{Hyperparameters for Atari 2600 datasets.}
     \label{tab:params_atari}
\end{longtable}

\begin{table*}[h]
  \centering
    \begin{tabular}{|l||S[table-format=3.1(2.1)]|S[table-format=3.1(3.1)]|S[table-format=4.1(3.1)]|S[table-format=4.1(3.1)]|}
     \hline
      \textbf{Dataset} & \textbf{NFQ}     & \textbf{QR-DQN}    & \textbf{BCQ}     & \textbf{CQL} \\
	 \hline
	 \hline
      Pong             & -18.0 \pm 0.9    & -16.6 \pm 1.1      & 13.8 \pm 1.5     & 16.3 \pm 1.4 \\
	 \hline
      Breakout         & 6.4 \pm 1.0      & 9.0 \pm 1.9        & 40.7 \pm 10.8    & 89.8 \pm 12.2 \\
	 \hline
      Qbert            & 648.5 \pm 81.4   & 695.8 \pm 153.1    & 8058.8 \pm 926.1 & 15791.5 \pm 530.5 \\
	 \hline
      Seaquest         & 841.2 \pm 67.3   & 603.2 \pm 62.3     & 1005.1 \pm 175.5 & 820.5 \pm 94.0 \\
	 \hline
      Asterix          & 745.0 \pm 85.6   & 551.0 \pm 76.2     & 857.0 \pm 61.2   & 1636.5 \pm 166.5 \\
	 \hline
  \end{tabular}
  \caption{The raw scores collected with the best trained policy in 1\% portion of Atari 2600 datasets. The scores are averaged over 10 random seeds.}
   \label{tab:benchmark_atari}
\end{table*}

\begin{table}[h]
  \centering
    \begin{tabular}{|l||S[table-format=3.1(3.1)]|r|}
     \hline
      \textbf{Dataset} & {d3rlpy}         & reference \\
	 \hline
	 \hline
      Pong             & -16.6 \pm 1.1    & -13.8      \\
	 \hline
     Breakout          & 9.0 \pm 1.9      & 7.9        \\
	 \hline
     Qbert             & 695.8 \pm 153.1  & 383.6        \\
	 \hline
     Seaquest          & 603.2 \pm 62.3   & 672.9        \\
	 \hline
     Asterix           & 551.0 \pm 76.2   & 166.3        \\
	 \hline
  \end{tabular}
    \caption{Side-by-side comparison with reference raw scores of QR-DQN reported in \cite{cql}, which only provides mean scores.}
  \label{tab:qr_dqn_ref}
\end{table}

\begin{table}[h]
  \centering
    \begin{tabular}{|l||S[table-format=5.1(3.1)]|r|}
     \hline
      \textbf{Dataset} & {d3rlpy}           & reference \\
	 \hline
	 \hline
      Pong             & 16.3 \pm 1.4       & 19.3      \\
	 \hline
     Breakout          & 89.8 \pm 12.2      & 61.1        \\
	 \hline
     Qbert             & 15791.5 \pm 530.5  & 14012.0        \\
	 \hline
     Seaquest          & 820.5 \pm 94.0     & 779.4        \\
	 \hline
     Asterix           & 1636,5 \pm 166.5   & 592.4        \\
	 \hline
  \end{tabular}
    \caption{Side-by-side comparison with reference raw scores of CQL reported in \cite{cql}, which only provides mean scores.}
  \label{tab:discrete_cql_ref}
\end{table}

\clearpage
\section*{Appendix C. Benchmark Results: Fine-tuning}
We evaluated the implemented algorithms with AntMaze datasets~\citep{d4rl} in a fine-tuning scenario where a policy is pretrained with a static dataset and fine-tuned with online experiences.
We followed the experimental procedure described in \cite{iql}.
In this evaluation, we chose AWAC and IQL, which are proposed as RL algorithms with fine-tuning capability.
Table~\ref{tab:antmaze_params} shows hyperparameters used in benchmarking.
All reward values are subtracted by $1$.
We used the same hyperparameters described in the original paper~\citep{awac,iql}.
In each training, the policy was fine-tuned for 1M steps after pretraining.
We repeated all experiments with 10 random seeds.

Table~\ref{tab:benchmark_antmaze} shows benchmark results.
Table~\ref{tab:finetune_awac_ref} and \ref{tab:finetune_iql_ref} show side-by-side comparisons with reference scores.
Considering that the authors did not report standard deviations, we believe that performance discrepancy is trivial.

\begin{small}
\begin{longtable}[c]{c|l|l}
    \hline
    Algorithm                                    & Hyperparameter            & Value \\
    \hline
    \multirow{7}{*}{AWAC~\citep{awac}}           & Critic learning rate      & 3e-4 \\
                                                 & Actor learning rate       & 3e-4 \\
                                                 & Mini-batch size           & 1024 \\
                                                 & $\lambda$~\citep{awac}    & 1 \\
                                                 & Actor hidden units        & [256, 256, 256, 256] \\
                                                 & Actor weight decay        & 1e-4 \\
                                                 & Critic hidden units       & [256, 256, 256, 256] \\
                                                 & Pretraining steps         & 25000 \\
    \hline
    \multirow{7}{*}{IQL~\citep{iql}}             & Critic learning rate      & 3e-4 \\
                                                 & Actor learning rate       & 3e-4 \\
                                                 & V-function learning rate  & 3e-4 \\
                                                 & Mini-batch size           & 256 \\
                                                 & Expectile                 & 0.9 \\
                                                 & Inverse temperature       & 10.0 \\
                                                 & Actor learning rate scheduler & Cosine \\
                                                 & Pretraining steps         & 1M \\
    \hline
    \caption{Hyperparameters for fine-tuning experiments.}
     \label{tab:antmaze_params}
\end{longtable}
\end{small}

\begin{table*}[h]
  \centering
    \begin{tabular}{|l||r|r|}
     \hline
      \textbf{Dataset}       & \textbf{AWAC}                                                                           & \textbf{IQL}    \\
	 \hline
	 \hline
        antmaze-umaze-v0       & 52.0$\pm$24.4  $\rightarrow$ 92.0$\pm$20.9                                            & 92.0$\pm$4.0 \hspace{0.04cm} $\rightarrow$ 98.0$\pm$4.0 \hspace{0.03cm}   \\
	 \hline
        antmaze-medium-play-v0 & 0.0$\pm$0.0 \hspace{0.07cm} $\rightarrow$ \hspace{0.08cm} 0.0$\pm$0.0 \hspace{0.08cm} & 71.0$\pm$14.5 $\rightarrow$ 91.0$\pm$9.4 \hspace{0.03cm}    \\
	 \hline
        antmaze-large-play-v0  & 0.0$\pm$0.0 \hspace{0.07cm} $\rightarrow$ \hspace{0.08cm} 0.0$\pm$0.0 \hspace{0.08cm} & 52.0$\pm$16.6 $\rightarrow$ 69.0$\pm$13.7 \\
	 \hline
  \end{tabular}
    \caption{The normalized scores collected with the final trained policy in AntMaze datasets~\citep{d4rl}. The numbers on the left represent scores of the pretrained policies. The numbers on the right represent scores of the fine-tuned policies. The scores are averaged over 10 random seeds.}
   \label{tab:benchmark_antmaze}
\end{table*}

\begin{table*}[h]
  \centering
  \begin{tabular}{|l||r|r|}
     \hline
      \textbf{Dataset}       & \textbf{d3rlpy}                                                                       & \textbf{reference}    \\
	 \hline
	 \hline
      antmaze-umaze-v0       & 52.0$\pm$24.4 $\rightarrow$ 92.0$\pm$20.9                                             & 56.7 $\rightarrow$ 59.0    \\
	 \hline
      antmaze-medium-play-v0 & 0.0$\pm$0.0 \hspace{0.07cm} $\rightarrow$ \hspace{0.08cm} 0.0$\pm$0.0 \hspace{0.08cm} & 0.0  $\rightarrow$ \hspace{0.08cm} 0.0    \\
	 \hline
      antmaze-large-play-v0  & 0.0$\pm$0.0 \hspace{0.07cm} $\rightarrow$ \hspace{0.08cm} 0.0$\pm$0.0 \hspace{0.08cm} & 0.0  $\rightarrow$ \hspace{0.08cm} 0.0 \\
	 \hline
  \end{tabular}
    \caption{Side-by-side comparison with reference normalized scores of AWAC reported in \cite{iql}, which only provides mean scores.}
   \label{tab:finetune_awac_ref}
\end{table*}

\begin{table*}[h]
  \centering
  \begin{tabular}{|l||r|r|}
     \hline
      \textbf{Dataset}       & \textbf{d3rlpy}                                                         & \textbf{reference}    \\
	 \hline
	 \hline
      antmaze-umaze-v0       & 92.0$\pm$4.0 \hspace{0.07cm} $\rightarrow$ 98.0$\pm$4.0 \hspace{0.08cm} & 86.7 $\rightarrow$ 96.0    \\
	 \hline
      antmaze-medium-play-v0 & 71.0$\pm$14.5 $\rightarrow$ 91.0$\pm$9.4 \hspace{0.08cm}                & 72.0 $\rightarrow$ 95.0    \\
	 \hline
      antmaze-large-play-v0  & 52.0$\pm$16.6 $\rightarrow$ 69.0$\pm$13.7                               & 25.5 $\rightarrow$ 46.0 \\
	 \hline
  \end{tabular}
    \caption{Side-by-side comparison with reference normalized scores of IQL reported in \cite{iql}, which only provides mean scores.}
   \label{tab:finetune_iql_ref}
\end{table*}

\bibliography{bibtex}

\end{document}